\documentclass[runningheads]{llncs}
\usepackage{graphicx}
\usepackage{amsmath,amssymb} 
\usepackage{color}
\usepackage{comment}
\usepackage[width=122mm,left=12mm,paperwidth=146mm,height=193mm,top=12mm,paperheight=217mm]{geometry}

\newcommand{\OURS}{3DMV}

\newcommand{\IGNORE}[1]{{}}
\newcommand{\ANGIE}[1]{\textcolor{blue}{{[Angie: #1]}}}
\newcommand{\MATTHIAS}[1]{\textcolor{red}{{\textbf{[Matthias: #1]}}}}

\def \path{\bp C}

\begin{document}
\pagestyle{headings}
\mainmatter
\def\ECCV18SubNumber{2207}  

\title{\OURS: Joint 3D-Multi-View Prediction for 3D Semantic Scene Segmentation} 

\titlerunning{\OURS: Joint 3D-Multi-View Prediction for 3D Semantic Scene Segmentation}
\authorrunning{A. Dai and M. Nie{\ss}ner}
\let\svthefootnote\thefootnote
\author{
	Angela Dai$^{1}$~~~~~~Matthias Nie{\ss}ner$^{2}$
}
\institute {
	$^{1}$Stanford University~~~$^{2}$Technical University of Munich
	\let\thefootnote\relax\footnote{\scriptsize{Corresponding author: \email{adai@cs.stanford.edu}}}\addtocounter{footnote}{-1}\let\thefootnote\svthefootnote
}

\maketitle
\begin{centering}
\begin{figure} 
	\centering
	\includegraphics[width=1.00\linewidth]{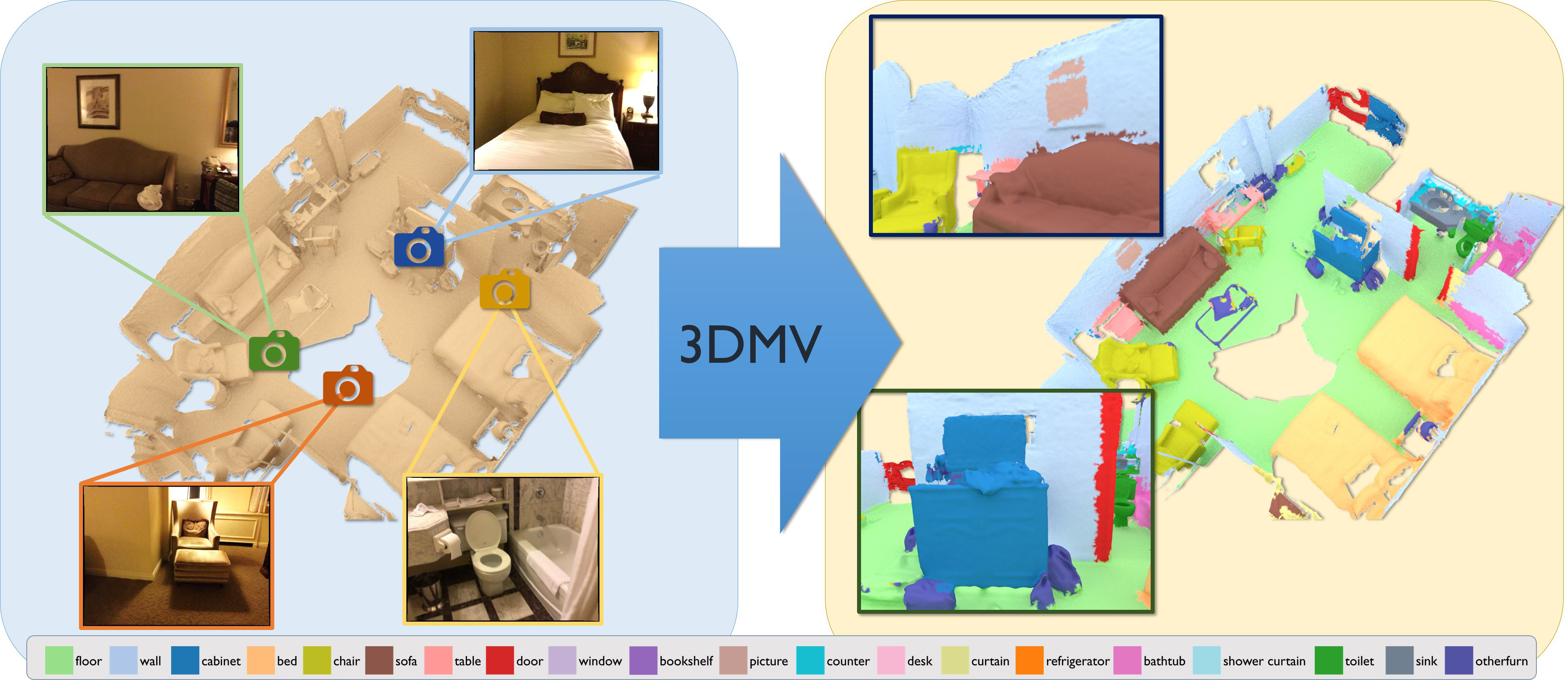}
	\caption{\OURS{} takes as input a reconstruction of an RGB-D scan along with its color images (left), and predicts a 3D semantic segmentation in the form of per-voxel labels (mapped to the mesh, right). The core of our approach is a joint 3D-multi-view prediction network that leverages the synergies between geometric and color features.
		\vspace{-0.4cm}}
	\label{fig:teaser}
\end{figure} 
\end{centering}



\begin{abstract}
We present \OURS{}, a novel method for 3D semantic scene segmentation of RGB-D scans  in indoor environments using a joint 3D-multi-view prediction network.
In contrast to existing methods that either use geometry {\em or} RGB data as input for this task, we combine both data modalities in a joint, end-to-end network architecture.
Rather than simply projecting color data into a volumetric grid and operating solely in 3D -- which would result in insufficient detail -- we first extract feature maps from associated RGB images.
These features are then mapped into the volumetric feature grid of a 3D network using a differentiable backprojection layer.
Since our target is 3D scanning scenarios with possibly many frames, we use a multi-view pooling approach in order to handle a varying number of RGB input views.
This learned combination of RGB and geometric features with our joint 2D-3D architecture achieves significantly better results than existing baselines.
For instance, our final result on the ScanNet 3D segmentation benchmark \cite{dai2017scannet} {\bf increases from 52.8\% to 75\% accuracy} compared to existing volumetric architectures.

\end{abstract}


\section{Introduction}
\label{sec:intro}

Semantic scene segmentation is important for a large variety of applications as it enables understanding of visual data.
In particular, deep learning-based approaches have led to remarkable results in this context, allowing prediction of accurate per-pixel labels in images \cite{long2015fully,he2017mask}.
Typically, these approaches operate on a single RGB image; however, one can easily formulate the analogous task in 3D on a per-voxel basis \cite{dai2017scannet}, which is a common scenario in the context of 3D scene reconstruction methods.
In contrast to the 2D task, the third dimension offers a unique opportunity as it not only predicts semantics, but also provides a spatial semantic map  of the scene content based on the underlying 3D representation.
This is particularly relevant for robotics applications since a robot relies not only on information of {\em what} is in a scene but also needs to know {\em where} things are.

In 3D, the representation of a scene is typically obtained from RGB-D surface reconstruction methods \cite{newcombe2011kinectfusion,niessner2013hashing,kahler2015very,dai2017bundlefusion} which often store scanned geometry in a 3D voxel grid where the surface is encoded by an implicit surface function such as a signed distance field \cite{curless1996volumetric}.
One approach towards analyzing these reconstructions is to leverage a CNN with 3D convolutions, which has been used for shape classification \cite{wu20153d,qi2016volumetric}, and recently also for predicting dense semantic 3D voxel maps \cite{song2017ssc,dai2017scannet,dai2018scancomplete}.
In theory, one could simply add an additional color channel to the voxel grid in order to incorporate RGB information; however, the limited voxel resolution prevents encoding feature-rich image data.
%
%
%
%

In this work, we specifically address this problem of how to incorporate RGB information for the 3D semantic segmentation task, and leverage the combined geometric and RGB signal in a joint, end-to-end approach.
To this end, we propose a novel network architecture that takes as input the 3D scene representation as well as the input of nearby views in order to predict a dense semantic label set on the voxel grid.
Instead of mapping color data directly on the voxel grid, the core idea is to first extract 2D feature maps from 2D images using the full-resolution RGB input.
These features are then downsampled through convolutions in the 2D domain, and the resulting 2D feature map is subsequently backprojected into 3D space.
In 3D, we leverage a 3D convolutional network architecture to learn from both the backprojected 2D features as well as 3D geometric features.
This way, we can join the benefits of existing approaches and leverage all available information, significantly improving on existing approaches.

Our main contribution is the formulation of a joint, end-to-end convolutional neural network which learns to infer 3D semantics from both 3D geometry and 2D RGB input.
%
%
In our evaluation, we provide a comprehensive analysis of the design choices of the joint 2D-3D architecture, and compare it with current state of the art methods.
In the end, our approach increases 3D segmentation accuracy from 52.8\% to 75\% compared to the best existing volumetric architecture.

\section{Related Work}
\label{sec:relatedWork}

\paragraph{Deep Learning in 3D.}

An important avenue for 3D scene understanding has been opened through recent advances in deep learning.
Similar to the 2D domain, convolutional neural networks (CNNs) can operate in volumetric domains using an additional spatial dimension for the filter banks.
3D ShapeNets~\cite{shapenet2015} was one of the first works in this context; they learn a 3D convolutional deep belief network from a shape database.
Several works have followed, using 3D CNNs for object classification~\cite{maturana2015voxnet,qi2016volumetric} or generative scene completion tasks \cite{dai2017complete,han2017complete,dai2018scancomplete}.
In order to address the memory and compute requirements, hierarchical 3D CNNs have been proposed to more efficiently represent and process 3D volumes \cite{riegler2017OctNet,wang2017cnn,riegler2017octnetfusion,tatarchenko2017octree,hane2017hierarchical,han2017complete}.
The spatial extent of a 3D CNN can also be increased with dilated convolutions \cite{yu2015multi}, which have been used to predict missing voxels and infer semantic labels \cite{song2017ssc}, or by using a fully-convolutional networks, in order to decouple the dimensions of training and test time \cite{dai2018scancomplete}.
Very recently, we have seen also network architectures that operate on an (unstructured) point-based representation \cite{qi2017pointnet,qi2017pointnet++}.

\paragraph{Multi-view Deep Networks.}

An alternative way of learning a classifier on 3D input is to render the geometry, run a 2D feature extractor, and combine the extracted features using max pooling.
The multi-view CNN approach by Su et al.~\cite{su2015multi} was one of the first to propose such an architecture for object classification.
However, since the output is a classification score, this architecture does not spatially correlate the accumulated 2D features.
Very recently, a multi-view network has been proposed for part-based mesh segmentation~\cite{kalogerakis20173d}.
Here, 2D confidence maps of each part label are projected on top of ShapeNet~\cite{shapenet2015} models, where a mesh-based CRF accumulates inputs of multiple images to predict the part labels on the mesh geometry.
%
%
This approach handles only relatively small label sets (e.g., 2-6 part labels), and its input is 2D renderings of the 3D meshes; i.e., the multi-view input is meant as a replacement input for 3D geometry.
Although these methods are not designed for 3D semantic segmentation, we consider them as the main inspiration for our multi-view component.

%

Multi-view networks have also been proposed in the context of stereo reconstruction.
For instance, Choi et al.~\cite{choy20163d} use an RNN to accumulate features from different views and Tulsiani et al.~\cite{tulsiani2017multi} propose an unsupervised approach that takes multi-view input to learn a latent 3D space for 3D reconstruction.
%
%
Another work in the context of stereo reconstruction was proposed by Kar et al.~\cite{kar2017learning}, which uses a sequence of 2D input views to reconstruct ShapeNet \cite{shapenet2015} models.
An alternative way to combine several input views with 3D, is by projecting colors directly into the voxels, maintaining one channel for each input view per voxel \cite{ji2017surfacenet}.
However, due to memory requirements, this becomes impractical for a large number of input views.

\paragraph{3D Semantic Segmentation.}

Semantic segmentation on 2D images is a popular task and has been heavily explored using cutting-edge neural network approaches \cite{long2015fully,he2017mask}.
The analog task can be formulated in 3D, where the goal is to predict semantic labels on a per-voxel level \cite{valentin2015semanticpaint,vineet2015incremental}.
Although this is a relatively recent task, it is extremely relevant to a large range of applications, in particular, robotics, where a spatial understanding of the inferred semantics is essential.
For the 3D semantic segmentation task, several datasets and benchmarks have recently been developed. 
The ScanNet \cite{dai2017scannet} dataset introduced a 3D semantic segmentation task on approx. 1.5k RGB-D scans and reconstructions obtained with a Structure Sensor.
It provides ground truth annotations for training, validation, and testing directly on the 3D reconstructions; it also includes approx. 2.5 mio RGB-D frames whose 2D annotations are derived using rendered 3D-to-2D projections.
Matterport3D~\cite{Matterport3D} is another recent dataset of about 90 building-scale scenes in the same spirit as ScanNet; it includes fewer RGB-D frames (approx. 194,400) but has more complete reconstructions.
%



\section{Overview}
\label{sec:overview}

The goal of our method is to predict a 3D semantic segmentation based on the input of commodity RGB-D scans.
More specifically, we want to infer semantic class labels on per-voxel level of the grid of a 3D reconstruction.
To this end, we propose a joint 2D-3D neural network that leverages both RGB and geometric information obtained from a 3D scans.
For the geometry, we consider a regular volumetric grid whose voxels encode a ternary state (known-occupied, known-free, unknown).
To perform semantic segmentation on full 3D scenes, our network operates on a per-chunk basis; i.e., it predicts columns of a scene in a sliding-window fashion through the $xy$-plane at test time.
For a given $xy$-location in a scene, the network takes as input the volumetric grid of the surrounding area (chunks of $31 \times 31 \times 62$ voxels).
The network then extracts geometric features using a series of 3D convolutions, and predicts per-voxel class labels for the center column at the current $xy$-location.
In addition to the geometry, we select nearby RGB views at the current $xy$-location that overlap with the associated chunk.
For all of these 2D views, we run the respective images through a 2D neural network that extracts their corresponding features.
Note that all of these 2D networks have the same architecture and share the same weights.

In order to combine the 2D and 3D features, we introduce a differentiable backprojection layer that maps 2D features onto the 3D grid.
These projected features are then merged with the 3D geometric information through a 3D convolutional part of the network.
In addition to the projection, we add a voxel pooling layer that enables handling a variable number of RGB views associated with a 3D chunk; the pooling is performed on a per-voxel basis. 
In order to run 3D semantic segmentation for entire scans, this network is run for each $xy$-location of a scene, taking as input the corresponding local chunks.

In the following, we will first introduce the details of our network architecture (see Sec.~\ref{sec:network}) and then show how we train and implement our method (see Sec.~\ref{sec:training}).

\section{Network Architecture}
\label{sec:network}

\begin{figure}[t!] \centering
	\includegraphics[width=0.95\linewidth]{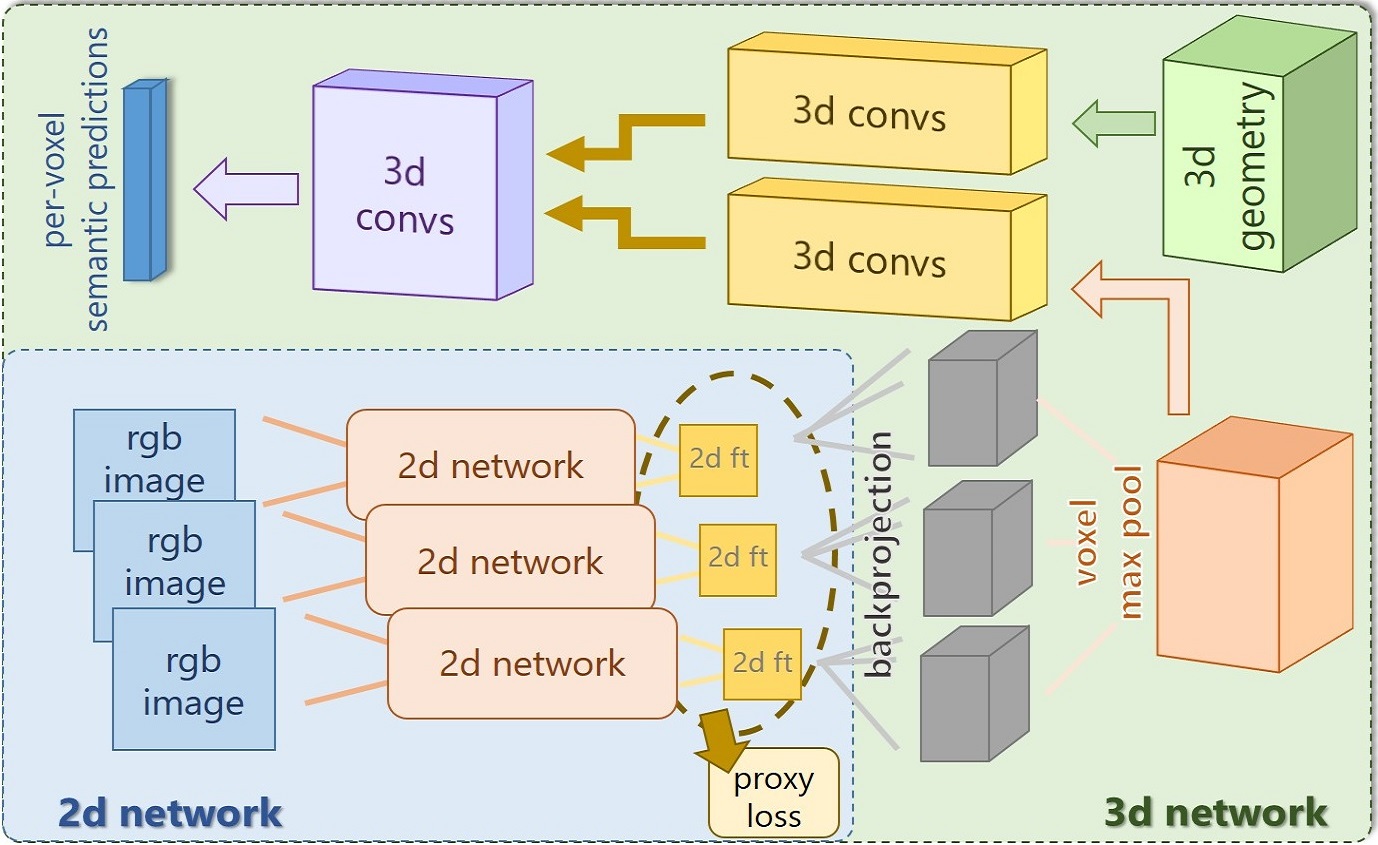}
	\caption{Network overview: our architecture is composed of a 2D and a 3D part.
	The 2D side takes as input several aligned RGB images from which features are learned with a proxy loss. 
	These are mapped to 3D space using a differentiable backprojection layer. 
	Features from multiple views are max-pooled on a per-voxel basis and fed into a stream of 3D convolutions. At the same time, we input the 3D geometry into another 3D convolution stream.
	Then, both 3D streams are joined and the 3D per-voxel labels are predicted. The whole network is trained in an end-to-end fashion.
	\vspace{-0.6cm}}
	\label{fig:network}
\end{figure}

Our network is composed of a 3D stream and several 2D streams that are combined in a joint 2D-3D network architecture.
The 3D part takes as input a volumetric grid representing the geometry of a 3D scan, and the 2D streams take as input the associated RGB images.
To this end, we assume that the 3D scan is composed of a sequence of  RGB-D images obtained from a commodity RGB-D camera, such as a Kinect or a Structure Sensor; although note that our method generalizes to other sensor types.
We further assume that the RGB-D images are aligned with respect to their world coordinate system using an RGB-D reconstruction framework; in the case of ScanNet~\cite{dai2017scannet} scenes, the BundleFusion \cite{dai2017bundlefusion} method is used.
Finally, the RGB-D images are fused together in a volumetric grid, which is commonly done by using an implicit signed distance function \cite{curless1996volumetric}.
An overview of the network architecture is provided in Fig.~\ref{fig:network}.
%

\subsection{3D Network}

Our 3D network part is composed of a series of 3D convolutions operating on a regular volumetric gird.
The volumetric grid is a subvolume of the voxelized 3D representation of the scene.
Each subvolume is centered around a specific $xy$-location at a size of $31\times 31\times 62$ voxels, with a voxel size of $4.8$cm.
Hence, we consider a spatial neighborhood of $1.5$m $\times$ $1.5$m and $3$m in height. 
Note that we use a height of $3$m in order to cover the height of most indoor environments, such that we only need to train the network to operate in varying $xy$-space.
The 3D network takes these subvolumes as input, and predicts the semantic labels for the center columns of the respective subvolume at a resolution of $1\times 1\times 62$ voxels; i.e., it simultaneously predicts labels for 62 voxels.
For each voxel, we encode the corresponding value of the scene reconstruction state: known-occupied (i.e., on the surface), known-free space (i.e., based on empty space carving \cite{curless1996volumetric}), or unknown space (i.e., we have no knowledge about the voxel).
We represent this through a 2-channel volumetric grid, the first a binary encoding of the occupancy, and the second a binary encoding of the known/unknown space.
The 3D network then processes these subvolumes with a series of nine 3D convolutions which expand the feature dimension and reduce the spatial dimensions, along with dropout regularization during training, before a final set of fully connected layers which predict the classification scores for each voxel.

In the following, we show how to incorporate learned 2D features from associated 2D RGB views.


\subsection{2D Network}

The aim of the 2D part of the network is to extract features from each of the input RGB images.
To this end, we use a 2D network architecture based on ENet~\cite{paszke2016enet} to learn those features.
Note that although we can use a variable of number of 2D input views, all 2D networks share the same weights as they are jointly trained.
Our choice to use ENet is due to its simplicity as it is both fast to run and memory-efficient to train.
In particular, the low memory requirements are critical since it allows us to jointly train our 2D-3D network in an end-to-end fashion with multiple input images per train sample. 
Although our aim is 2D-3D end-to-end training, we additionally use a 2D proxy loss for each image that allows us to make the training more stable; i.e., each 2D stream is asked to predict meaningful semantic features for an RGB image segmentation task.
Here, we use semantic labels of the 2D images as ground truth; in the case of ScanNet~\cite{dai2017scannet}, these are derived from the original 3D annotations by rendering the annotated 3D mesh from the camera points of the respective RGB image poses.
The final goal of the 2D network is to obtain the features in the last layer before the proxy loss per-pixel classification scores; these features maps are then backprojected into 3D to join with the 3D network, using a differentiable backprojection layer.
In particular, from an input RGB image of size $328\times 256$, we obtain a 2D feature map of size $(128\times) 41\times 32$, which is then backprojected into the space of the corresponding 3D volume, obtaining a 3D representation of the feature map of size $(128\times) 31\times 31\times 62$.

\subsection{Backprojection Layer}

In order to connect the learned 2D features from each of the input RGB views with the 3D network, we use a differentiable backprojection layer.
Since we assume known 6-DoF pose alignments for the input RGB images with respect to each other and the 3D reconstruction, we can compute 2D-3D associations on-the-fly.
The layer is essentially a loop over every voxel in 3D subvolume where a given image is associated to.
For every voxel, we compute the 3D-to-2D projection based on the corresponding camera pose, the camera intrinsics, and the world-to-grid transformation matrix.
We use the depth data from the RGB-D images in order to prune projected voxels beyond a threshold of the voxel size of $4.8$cm; i.e., we compute only associations for voxels close to the geometry of the depth map.
We compute the correspondences from 3D voxels to 2D pixels since this allows us to obtain a unique voxel-to-pixel mapping.
Although one could pre-compute these voxel-to-pixel associations, we simply compute this mapping on-the-fly in the layer as these computations are already highly memory bound on the GPU; in addition, it saves significant disk storage since this it would involve a large amount of index data for full scenes.

Once we have computed voxel-to-pixel correspondences, we can project the features of the last layer of the 2D network to the voxel grid:
$$n_{feat}\times w_{2d} \times h_{2d} \rightarrow n_{feat}\times w_{3d} \times h_{3d} \times d_{3d}$$
For the backward pass, we use the inverse mapping of the forward pass, which we store in a temporary index map.
We use 2D feature maps (feature dim. of $128$) of size $(128 \times) 41 \times 31$ and project them to a grid of size $(128\times) 31\times 31\times 62$.

In order to handle multiple 2D input streams, we compute voxel-to-pixel associations with respect to each input view.
As a result, some voxels will be associated with multiple pixels from different views.
In order to combine projected features from multiple input views, we use a voxel max-pooling operation that computes the maximum response on a per feature channel basis.
Since the max pooling operation is invariant to the number of inputs, it enables selecting for the features of interest from an arbitrary number of input images.


\subsection{Joint 2D-3D Network}

The joint 2D-3D network combines 2D RGB features and 3D geometric features using the mapping from the backprojection layer.
These two inputs are processed with a series of 3D convolutions, and then concatenated together; the joined feature is then further processed with a set of 3D convolutions.
We have experimented with several options as to where to join these two parts: at the beginning (i.e., directly concatenated together without independent 3D processing), approximately 1/3 or 2/3 through the 3D network, and at the end (i.e., directly before the classifier). 
We use the variant that provided the best results, fusing the 2D and 3D features together at 2/3 of the architectures (i.e., after the 6th 3D convolution of 9); see Tab.~\ref{table:combining2d3d} for the corresponding ablation study.
%
%
Note that the entire network, as shown in Fig.~\ref{fig:network}, is trained in an end-to-end fashion, which is feasible since all components are differentiable.
Tab.~\ref{table:learnable_weights} shows an overview of the distribution of learnable parameters of our \OURS{} model.

\begin{table*}[!htb]
	\centering
	\begin{tabular}{|c|c|c|c|c|}
		\hline
		& \ \ \ 2d only \ \ \ & 3d (2d ft only) & 3d (3d geo only) & 3d (fused 2d/3d)\\ \hline
		\# trainable params & 146,176 & 379,744 & 87,136 & 10,224,300\\ \hline
	\end{tabular}
	\caption{Distribution of learnable parameters of our \OURS{} model. Note that the majority of the network weights are part of the combined 3D stream just before the per-voxel predictions where we rely on strong feature maps; see top left of Fig.~\ref{fig:network}.
		\vspace{-0.5cm}}
	\label{table:learnable_weights}
\end{table*}


\subsection{Evaluation in Sliding Window Mode}

Our joint 2D-3D network operates on a per-chunk basis; i.e., it takes fixed subvolumes of a 3D scene as input (along with associated RGB views), and predicts labels for the voxels in the center column of the given chunk.
In order to perform a semantic segmentation of large 3D environments, we slide the subvolume through the 3D grid of the underlying reconstruction.
Since the height of the subvolume ($3$m) is sufficient for most indoor environments, we only need to slide over the $xy$-domain of the scene.
Note, however, that for training, the training samples do not need to be spatially connected, which allows us to train on a random set of subvolumes.
This de-coupling of training and test extents is particularly important since it allows us to provide a good label and data distribution of training samples (e.g., chunks with sufficient coverage and variety).



\section{Training}
\label{sec:training}

\subsection{Training Data}

We train our joint 2D-3D network architecture in an end-to-end fashion.
To this end, we prepare correlated 3D and RGB input to the network for the training process.
The 3D geometry is encoded in a ternary occupancy grid that encodes known-occupied, known-free, and unknown states for each voxel.
The ternary information is split upon 2 channels, where the first channel encodes occupancy and the second channel encodes the known vs. unknown state.
To select train subvolumes from a 3D scene, we randomly sample subvolumes as potential training samples.
For each potential train sample, we check its label distribution and discard samples containing only structural elements (i.e., wall/floor) with $95\%$ probability.
In addition, all samples with empty center columns are discarded as well as samples with less than 70\% of the center column geometry annotated.

For each subvolume, we then associate $k$ nearby RGB images whose alignment is known from the 6-DoF camera pose information.
We select images greedily based on maximum coverage; i.e., we first pick the image covering the most voxels in the subvolume, and subsequently take each next image which covers the most number of voxels not covered by current set. 
We typically select 3-5 images since additional gains in coverage become smaller with each added image. 
%
For each sampled subvolume, we augment it with 8 random rotations for a total of $1,316,080$ train samples.
Since existing 3D datasets, such as ScanNet \cite{dai2017scannet} or Matterport3D \cite{Matterport3D} contain unannotated regions in the ground truth (see Fig.~\ref{fig:comparison_scannet}, right), we mask out these regions in both our 3D loss and 2D proxy loss.
Note that this strategy still allows for making predictions for all voxels at test time.



\subsection{Implementation}

We implement our approach in PyTorch.
While 2D and 3D conv layers are already provided by the PyTorch API, we implement a custom layer for the backprojection layer.
We implement this backprojection in python, as a custom PyTorch layer, representing the projection as series of matrix multiplications in order to exploit PyTorch parallelization, and run the backprojection on the GPU through the PyTorch API.
%
For training, we have tried only training parts of the network; however, we found that the end-to-end version that jointly optimizes both 2D and 3D performed best.
In the training processes, we use an SGD optimizer with a learning rate of $0.001$ and a momentum of $0.9$; we set the batch size to $8$.
Note that our training set is quite biased towards structural classes (e.g., wall, floor), even when discarding most structural-only samples, as these elements are vastly dominant in indoor scenes.
In order to account for this data imbalance, we use the histogram of classes represented in the train set to weight the loss during training.
We train our network for $200,000$ iterations; for our network trained on $3$ views, this takes $\approx 24$ hours, and for $5$ views, $\approx 48$ hours.



\section{Results}
\label{sec:results}

\begin{figure}[t!] \centering
	\includegraphics[width=0.92\linewidth]{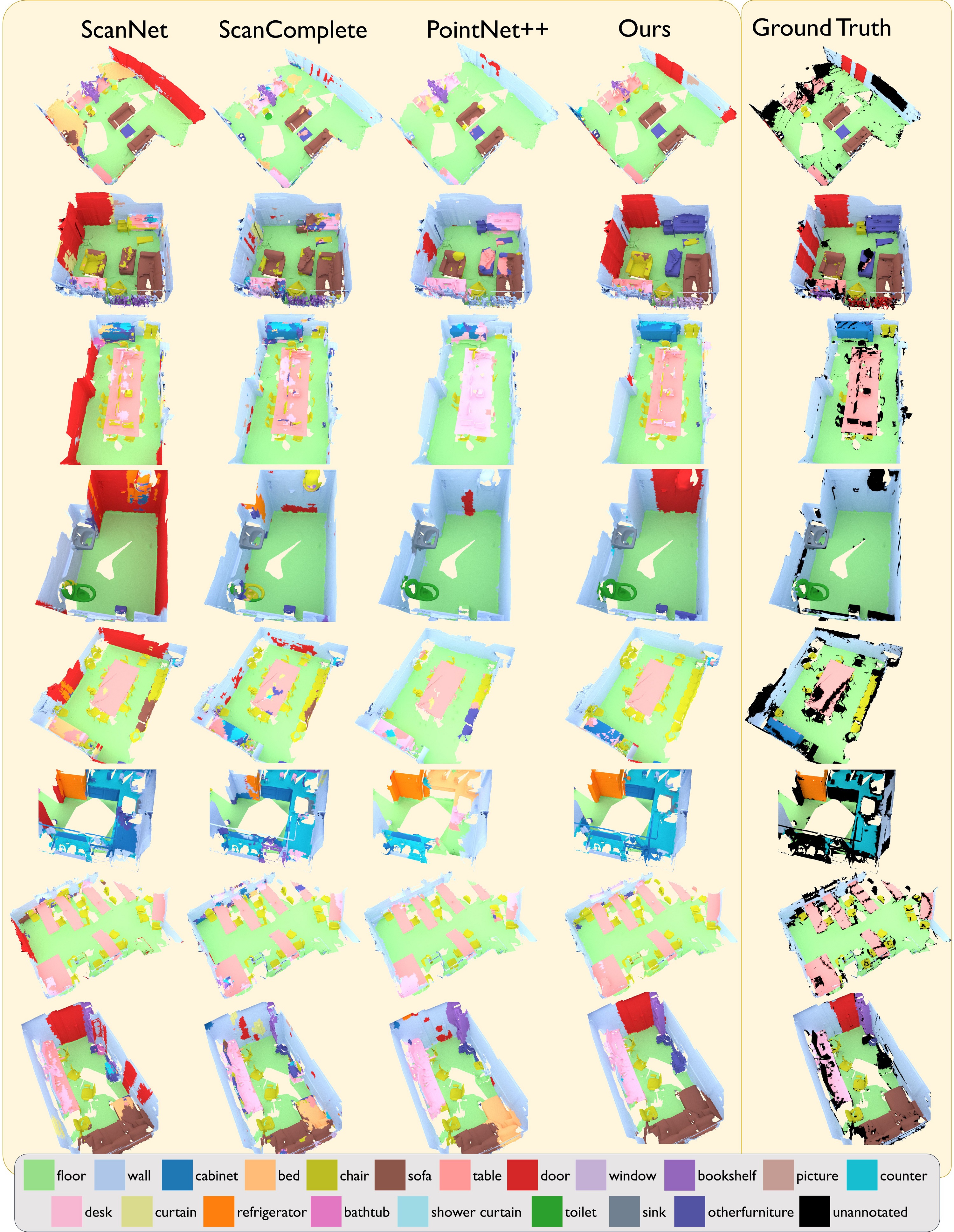}
	\caption{Qualitative semantic segmentation results on the ScanNet~\cite{dai2017scannet} test set. We compare with the 3D-based approaches of ScanNet~\cite{dai2017scannet}, ScanComplete~\cite{dai2018scancomplete}, PointNet++~\cite{qi2017pointnet++}. Note that the ground truth scenes contain some unannotated regions, denoted in black. Our joint 3D-multi-view approach achieves more accurate semantic predictions.
		\vspace{-0.4cm}}
	\label{fig:comparison_scannet}
\end{figure}

\begin{figure}[t!] \centering
	\includegraphics[width=0.95\linewidth]{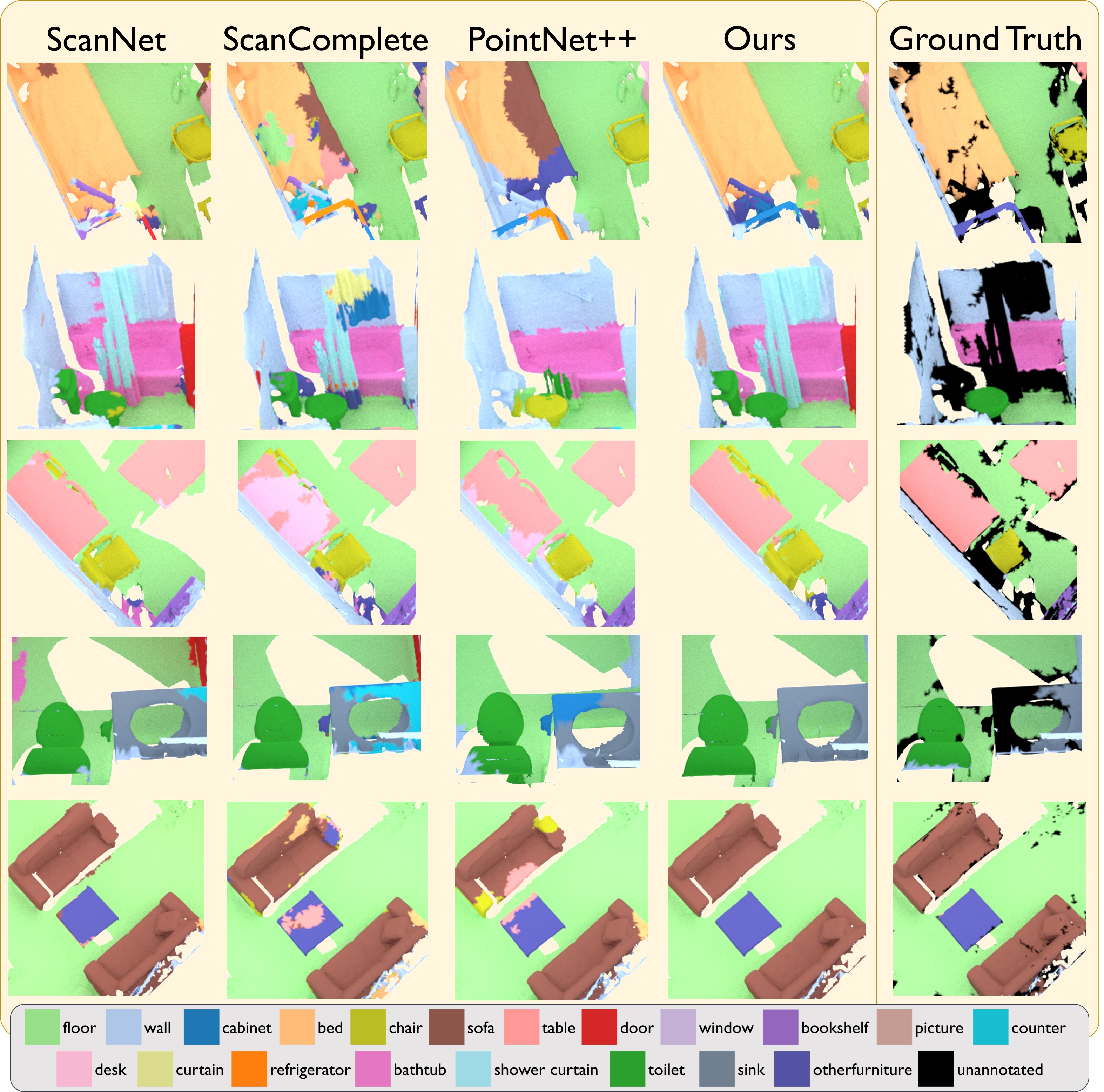}
	\caption{Additional qualitative semantic segmentation results (close ups) on the ScanNet~\cite{dai2017scannet} test set. Note the consistency of our predictions compared to the other baselines.
		\vspace{-0.4cm}}
	\label{fig:comparison_scannet_zoom}
\end{figure}

In this section, we provide an evaluation of our proposed method with a comparison to existing approaches.
%
We evaluate on the ScanNet dataset \cite{dai2017scannet}, which contains 1513 RGB-D scans composed of 2.5M RGB-D images.
We use the public train/val/test split of 1045, 156, 312 scenes, respectively, and follow the 20-class semantic segmentation task defined in the original ScanNet benchmark.
We evaluate our results with per-voxel class accuracies, following the evaluations of previous work~\cite{dai2017scannet,qi2017pointnet++,dai2018scancomplete}.
Additionally, we visualize our results qualitatively and in comparison to previous work in Fig~\ref{fig:comparison_scannet}, with close-ups shown in Fig~\ref{fig:comparison_scannet_zoom}. Note that we map the predictions from all methods back onto the mesh reconstruction for ease of visualization.

\vspace{0.2cm}\noindent
\textbf{Comparison to state of the art.}
Our main results are shown in Tab.~\ref{table:comparison_scannet}, where we compare to several state-of-the-art volumetric (ScanNet\cite{dai2017scannet}, ScanComplete\cite{dai2018scancomplete}) and point-based approaches (PointNet++\cite{qi2017pointnet++}) on the ScanNet test set.
Additionally, we show an ablation study regarding our design choices in Tab.~\ref{table:ablation_scannet}.
%

The best variant of our \OURS{} network achieves 75\% average classification accuracy which is quite significant considering the difficulty of the task and the performance of existing approaches.
That is, we improve 22.2\% over existing volumetric and 14.8\% over the state-of-the-art PointNet++ architecture.

\vspace{0.2cm}\noindent
\textbf{How much does RGB input help?}
Tab.~\ref{table:ablation_scannet} includes a direct comparison between our 3D network architecture when using RGB features against the exact same 3D network without the RGB input.
Performance improves from 54.4\% to 70.1\% with RGB input, even with just a single RGB view.
In addition, we tried out the naive alternative of using per-voxel colors rather than a 2D feature extractor.
Here, we see only a marginal difference compared to the purely geometric baseline (54.4\% vs. 55.9\%).
We attribute this relatively small gain to the limited grid resolution ($\approx 5$cm voxels), which is insufficient to capture rich RGB features.
Overall, we can clearly see the benefits of RGB input, as well as the design choice to first extract features in the 2D domain.

\vspace{0.2cm}\noindent
\textbf{How much does geometric input help?}
Another important question is whether we actually need the 3D geometric input, or whether geometric information is a redundant subset of the RGB input; see Tab.~\ref{table:ablation_scannet}.
The first experiment we conduct in this context is simply a projection of the predicted 2D labels on top of the geometry.
If we only use the labels from a single RGB view, we obtain 27\%  average accuracy (vs. 70.1\% with 1 view + geometry); for 3 views, this label backprojection achieves 44.2\% (vs. 73.0\% with 3 views + geometry). 
Note that this is related to the limited coverage of the RGB backprojections (see Tab.~\ref{table:coverage}).

However, the interesting experiment now is what happens if we still run a series of 3D convolutions after the backprojection of the 2D labels.
Again, we omit inputting the scene geometry, but we now learn how to combine and propagate the backprojected features in the 3D grid; essentially, we ignore the first part of our 3D network; cf.~Fig.~\ref{fig:network}.
For 3 RGB views, this results in an accuracy of 58.2\%; this is higher than the 54.4\% of geometry only; however, it is much lower than our final 3-view result of 73.0\% from the joint network.
Overall, this shows that the combination of RGB and geometric information aptly complements each other, and that the synergies allow for an improvement over the individual inputs by 14.8\% and 18.6\%, respectively (for 3 views).

\begin{table*}[!htb]
	\resizebox{\textwidth}{!}{%
		\centering
		\begin{tabular}{|l||c|c|c|c|c|c|c|c|c|c|c|c|c|c|c|c|c|c|c|c||c|}
			\hline
			& wall & floor & cab & bed & chair & sofa & table & door & wind & bkshf & pic & cntr & desk & curt & fridg & show & toil & sink & bath & other & avg \\ \hline
			ScanNet~\cite{dai2017scannet} & 70.1 & 90.3 & 49.8 & 62.4 &  69.3 & 75.7 & \textbf{68.4} & 48.9 & 20.1 & 64.6 & 3.4 & 32.1 & 36.8 & 7.0 & 66.4 & 46.8 & 69.9 & 39.4 & 74.3 & 19.5 &  50.8 \\ \hline
			ScanComplete~\cite{dai2018scancomplete} & 87.2 & 96.9 & 44.5 & 65.7 & 75.1 & 72.1 & 63.8 & 13.6 & 16.9 & 70.5 & 10.4 & 31.4 & 40.9 & 49.8 & 38.7 & 46.8 & 72.2 & 47.4 & 85.1 & 26.9  & 52.8 \\ \hline 
			PointNet++~\cite{qi2017pointnet++} & \textbf{89.5} & \textbf{97.8} & 39.8 & 69.7 & \textbf{86.0} & 68.3 & 59.6 & 27.5 & 23.7 & 84.3 & 0.0 & \textbf{37.6} & 66.7 & 48.7 & 54.7 & \textbf{85.0} & 84.8 & 62.8 & 86.1 & 30.7 & 60.2 \\ 
			\hline \hline
			\textbf{\OURS{} (ours)} & 73.9 & 95.6 & \textbf{69.9} & \textbf{80.7} &  85.9 & \textbf{75.8} & 67.8 &  \textbf{86.6} & \textbf{61.2} & \textbf{88.1} & \textbf{55.8} & 31.9 & \textbf{73.2} & \textbf{82.4} & \textbf{74.8} & 82.6 & \textbf{88.3} & \textbf{72.8} & \textbf{94.7} & \textbf{58.5} & \textbf{75.0}		\\ \hline
		\end{tabular}
	}
	\vspace{0.1cm}
	\caption{Comparison of our final trained model (5 views, end-to-end) against other state-of-the-art methods on the ScanNet dataset \cite{dai2017scannet}. We can see that our approach makes significant improvements, 22.2\% over existing volumetric and approx. 14.8\% over state-of-the-art PointNet++ architectures. 
		\vspace{-1.0cm}}
	\label{table:comparison_scannet}
\end{table*}

\vspace{0.2cm}\noindent
\textbf{How to feed 2D features into the 3D network?}
An interesting question is where to join 2D and 3D features; i.e., at which layer of the 3D network do we fuse together the features originating from the RGB images with the features from the 3D geometry.
On the one hand, one could argue that it makes more sense to feed the 2D part early into the 3D network in order to have more capacity for learning the joint 2D-3D combination.
On the other hand, it might make more sense to keep the two streams separate for as long as possible to first extract strong independent features before combining them.

To this end, we conduct an experiment with different 2D-3D network combinations (for simplicity, always using a single RGB view without end-to-end training); see Tab.~\ref{table:combining2d3d}.
We tried four combinations, where we fused the 2D and 3D features at the beginning, after the first third of the network, after the second third, and at the very end into the 3D network.
Interestingly, the results are relatively similar ranging from 67.6\%, 65.4\% to 69.1\% and 67.5\% suggesting that the 3D network can adapt quite well to the 2D features. 
Across these experiments, the second third option turned out to be a few percentage points higher than the alternatives; hence, we use that as a default in all other experiments.


\vspace{0.2cm}\noindent
\textbf{How much do additional views help?}
In Tab.~\ref{table:ablation_scannet}, we also examine the effect of each additional view on classification performance.
For geometry only, we obtain an average classification accuracy of 54.4\%; adding only a single view per chunk increases to 70.1\% (+15.7\%); for 3 views, it increases to 73.1\% (+3.0\%); for 5 views, it reaches 75.0\% (+1.9\%).
Hence, for every additional view the incremental gains become smaller; this is somewhat expected as a large part of the benefits are attributed to additional coverage of the 3D volume with 2D features.
If we already use a substantial number of views, each additional added feature shares redundancy with previous views, as shown in Tab.~\ref{table:coverage}.

\vspace{0.2cm}\noindent
\textbf{Is end-to-end training of the joint 2D-3D network useful?}
Here, we examine the benefits of training the 2D-3D network in an end-to-end fashion, rather than simply using a pre-trained 2D network.
We conduct this experiment with 1, 3, and 5 views.
The end-to-end variant consistently outperforms the fixed version, improving the respective accuracies by 1.0\%, 0.2\%, and 0.5\%.
Although the end-to-end variants are strictly better, the increments are smaller than we initially hoped for.
We also tried removing the 2D proxy loss that enforces good 2D predictions, which led to a slightly lower performance.
Overall, end-to-end training with a proxy loss always performed best and we use it as our default.


\vspace{-0.5cm}

\begin{table*}[!htb]
	\resizebox{\textwidth}{!}{%
		\centering
		\begin{tabular}{|l||c|c|c|c|c|c|c|c|c|c|c|c|c|c|c|c|c|c|c|c||c|}
			\hline
			& wall & floor & cab & bed & chair & sofa & table & door & wind & bkshf & pic & cntr & desk & curt & fridg & show & toil & sink & bath & other & avg \\ \hline
			2d only (1 view) & 37.1 & 39.1 & 26.7 & 33.1 & 22.7 & 38.8 & 17.5 & 38.7 & 13.5 & 32.6 & 14.9 & 7.8 & 19.1 & 34.4 & 33.2 & 13.3 & 32.7 & 29.2 & 36.3 & 20.4 & 27.1 \\ \hline
			2d only (3 views) & 58.6 & 62.5 & 40.8 & 51.6 & 38.6 & 59.7 & 31.1 & 55.9 & 25.9 & 52.9 & 25.1 & 14.2 & 35.0 & 51.2 & 57.3 & 36.0 & 47.1 & 44.7 & 61.5 & 34.3 & 44.2 \\ \hline
			Ours (no geo input) & 76.2 & 92.9 & 59.3 & 65.6 & 80.6 & 73.9 & 63.3 & 75.1 & 22.6 & 80.2 & 13.3 & 31.8 & 43.4 & 56.5 & 53.4 & 43.2 & 82.1 & 55.0 & 80.8 & 9.3 & 58.2 \\ \hline
			Ours (3d geo only) & 60.4 & 95.0 & 54.4 & 69.5 & 79.5 & 70.6 & 71.3 & 65.9 & 20.7 & 71.4 & 4.2 & 20.0 & 38.5 & 15.2 & 59.9 & 57.3 & 78.7 & 48.8 & 87.0 & 20.6 & 54.4 \\ \hline 
			Ours (3d geo+voxel color) & 58.8 & 94.7 & 55.5 & 64.3 & 72.1 & 80.1 & 65.5 & \textbf{70.7} & 33.1 & 69.0 & 2.9 & 31.2 & 49.5 & 37.2 & 49.1 & 54.1 & 75.9 & 48.4 & 85.4 & 20.5 & 55.9 \\ \hline 
			Ours (1 view, fixed 2d) & 77.3 & 96.8 & \textbf{70.0} & 78.2 & 82.6 & \textbf{85.0} & 68.5 & 88.8 & 36.0 & 82.8 & 15.7 & 32.6 & 60.3 & 71.0 & 76.7 & 82.2 & 74.8 & 57.6 & 87.0 & 58.5 & 69.1 \\ \hline
			Ours (1 view) & 70.7 & 96.8 & 61.4 & 76.4 & 84.4 & 80.3 & 70.4 & 83.9 & 57.9 & 85.3 & 41.7 & 35.0 & 64.5 & 75.6 & 81.3 & 58.2 & 85.0 & 60.5 & 81.6 & 51.7 & 70.1 \\ \hline
			Ours (3 view, fixed 2d) & \textbf{81.1} & 96.4 & 58.0 & 77.3 & 84.7 & 85.2 & \textbf{74.9} & 87.3 & 51.2 & 86.3 & 33.5 & \textbf{47.0} & 52.4 & 79.5 & 79.0 & 72.3 & 80.8 & \textbf{76.1} & 92.5 & \textbf{60.7} & 72.8 \\ \hline
			Ours (3 view) & 75.2 & \textbf{97.1} & 66.4 & 77.6 & 80.6 & 84.5 & 66.5 & 85.8 & 61.8 & 87.1 & 47.6 & 24.7 & 68.2 & 75.2 & 78.9 & 73.6 & 86.9 & \textbf{76.1} & 89.9 & 57.2 & 73.0 \\ \hline
			Ours (5 view, fixed 2d) & 77.3 & 95.7 & 68.9 & \textbf{81.7} & \textbf{89.6} & 84.2 & 74.8 & 83.1 & \textbf{62.0} & 87.4 & 36.0 & 40.5 & 55.9 & \textbf{83.1} & \textbf{81.6} & 77.0 & 87.8 & 70.7 & 93.5 & 59.6 & 74.5 \\ \hline
			\textbf{Ours (5 view)} & 73.9 & 95.6 & 69.9 & 80.7 &  85.9 & 75.8 & 67.8 &  86.6 & 61.2 & \textbf{88.1} & \textbf{55.8} & 31.9 & \textbf{73.2} & 82.4 & 74.8 & \textbf{82.6} & \textbf{88.3} & 72.8 & \textbf{94.7} & 58.5 & \textbf{75.0}		\\ \hline
		\end{tabular}
	}
	\vspace{0.1cm}
	\caption{Ablation study for different design choices of our approach on ScanNet \cite{dai2017scannet}. We first test simple baselines where we backproject 2D labels from 1 and 3 views (rows 1-2), then run set of 3D convs after the backprojections (row 3). We then test a 3D-geometry-only network (row 4). Augmenting the 3D-only version with per-voxel colors shows only small gains (row 5). In rows 6-11, we test our joint 2D-3D architecture with varying number of views, and the effect of end-to-end training. Our 5-view, end-to-end variant performs best. 
		\vspace{-0.5cm}}
	\label{table:ablation_scannet}
\end{table*}

\vspace{0.2cm}\noindent
\textbf{Evaluation in 2D domains using NYUv2.}
Although we are predicting 3D per-voxel labels, we can also project the obtained voxel labels into the 2D images.
In Tab.~\ref{table:denselabeling_nyu}, we show such an evaluation on the NYUv2~\cite{silberman11indoor} dataset.
For this task, we train our network on both ScanNet data as well as the NYUv2 train annotations projected into 3D.
%
Although this is not the actual task of our method, it can be seen as an efficient way to accumulate semantic information from multiple RGB-D frames by using the 3D geometry as a proxy for the learning framework.
Overall, our joint 2D-3D architecture compares favorably against the respective baselines on this 13-class task.

\begin{table*}[!htb]
		\centering
		\begin{tabular}{|c|c|c|c|}
			\hline
			 & 1 view & 3 views & 5 views \\ \hline
			coverage & 40.3\% & 64.4\% & 72.3\% \\ \hline
		\end{tabular}
	\caption{Amount of coverage from varying number of views over the annotated ground truth voxels of the ScanNet~\cite{dai2017scannet} test scenes.
		\vspace{-0.4cm}}
	\label{table:coverage}
\end{table*}

\begin{table*}[!htb]
	\resizebox{\textwidth}{!}{%
	\centering
	\begin{tabular}{|l||c|c|c|c|c|c|c|c|c|c|c|c|c|c|c|c|c|c|c|c||c|}
		\hline
		& wall & floor & cab & bed & chair & sofa & table & door & wind & bkshf & pic & cntr & desk & curt & fridg & show & toil & sink & bath & other & avg \\ \hline
		begin & 78.8 & 96.3 & 63.7 & 72.8  & 83.3 & 81.9 & \textbf{74.5} & 81.6 & 39.5 & 89.6 & \textbf{24.8} & 33.9 & 52.6 & \textbf{74.8} & 76.0 & 47.5 & 80.1 & \textbf{65.4} & 85.9 & 49.4 & 67.6 \\ \hline
		1/3 & 79.3 & 95.5 & 65.1 & 75.2 & 80.3 & 81.5 & 73.8 & 86.0 & 30.5 & \textbf{91.7} & 11.3 & \textbf{35.5} & 46.4 & 66.6 & 67.9 & 44.1 & \textbf{81.7}  & 55.5 & 85.9  & 53.3 & 65.4 \\ \hline
		\textbf{2/3} & 77.3 & \textbf{96.8} & \textbf{70.0} & \textbf{78.2} & 82.6 & \textbf{85.0} & 68.5 & \textbf{88.8} & 36.0 & 82.8 & 15.7 & 32.6 & \textbf{60.3} & 71.0 & \textbf{76.7} & \textbf{82.2} & 74.8 & 57.6 & 87.0 & \textbf{58.5} & \textbf{69.1} \\ \hline
		end & \textbf{82.7} & 96.3 & 67.1 & 77.8 & \textbf{83.2} & 80.1 & 66.0 & 80.3 & \textbf{41.0} & 83.9 & 24.3 & 32.4 & 57.7 & 70.1 & 71.5 & 58.5 & 79.6 & 65.1 & \textbf{87.2} & 45.8 & 67.5 \\ \hline
	\end{tabular}
}
	\vspace{0.1cm}
	\caption{Evaluation of various network combinations for joining the 2D and 3D streams in the 3D architecture (cf.~Fig.~\ref{fig:network}, top). We use the single view variant with a fixed 2D network here for simplicity. Interestingly, performance only changes slightly; however, the 2/3 version performed the best, which is our default for all other experiments.
		\vspace{-0.4cm}}
	\label{table:combining2d3d}

\end{table*}

\begin{table*}[!htb]
	\resizebox{\textwidth}{!}{%
	\centering
	\begin{tabular}{|l||c|c|c|c|c|c|c|c|c|c|c|c|c||c|}
		\hline
		& bed & books & ceil. & chair & floor & furn. & obj. & pic. & sofa & table & tv & wall & wind. & avg. \\ \hline
		SceneNet~\cite{handa2015scenenet} & 70.8 & 5.5 & 76.2 & 59.6 & 95.9 & 62.3 & 50.0 & 18.0 & 61.3 & 42.2 & 22.2 & 86.1 & 32.1 &  52.5 \\ \hline
		Hermans et al.~\cite{hermans2014dense} & 68.4 & 45.4 & 83.4 & 41.9 & 91.5 & 37.1 & 8.6 & 35.8 & 58.5 & 27.7 & 38.4 & 71.8 & 48.0 & 54.3 \\ \hline
		SemanticFusion~\cite{mccormac2017semanticfusion} (RGBD+CRF) & 62.0 & \textbf{58.4} & 43.3 & 59.5 & 92.7 & 64.4 & \textbf{58.3} & 65.8 & 48.7 & 34.3 & 34.3 & 86.3 & 62.3 & 59.2 \\ \hline
		SemanticFusion~\cite{mccormac2017semanticfusion,eigen2015predicting} (Eigen+CRF) & 48.3 & 51.5 & 79.0 & 74.7 & 90.8 & 63.5 & 46.9 & 63.6  & 46.5 & 45.9 & \textbf{71.5} & \textbf{89.4} & 55.6 & 63.6 \\ \hline
		ScanNet~\cite{dai2017scannet} & 81.4 & - & 46.2 & 67.6 &  \textbf{99.0} & 65.6 & 34.6 &  - & 67.2 & 50.9 & 35.8 & 55.8 & 63.1 & 60.7 \\ \hline \hline
		\textbf{\OURS{} (ours)} & \textbf{84.3} & 44.0 &  43.4 & \textbf{77.4} &  92.5 & \textbf{76.8} & 54.6 & \textbf{70.5} & \textbf{86.3} & \textbf{58.6} & 67.3 & 84.5 & \textbf{85.3} & \textbf{71.2}		\\ \hline
	\end{tabular}
}
	\vspace{0.1cm}
	\caption{We can also evaluate our method on 2D semantic segmentation tasks by projecting the predicted 3D labels into the respective RGB-D frames.
		Here, we show a comparison on dense pixel classification accuracy on NYU2~\cite{Silberman:ECCV12}. Note that the reported ScanNet classification is on the $11$-class task.
		\vspace{-0.5cm}}
	\label{table:denselabeling_nyu}
\end{table*}

\vspace{0.2cm}\noindent
\textbf{Summary Evaluation.}
\vspace{-0.2cm}
\begin{itemize}
	\item RGB and geometric features are orthogonal and help each other
	\item More views help, but increments get smaller with every view
	\item End-to-end training is strictly better, but the improvement is not that big.
	\item Variations of where to join the 2D and 3D features change performance to some degree; 2/3 performed best in our tests.
	\item Our results are significantly better than the best volumetric or PointNet baseline (+22.2\% and +14.8\%, respectively).
\end{itemize}

\noindent
\textbf{Limitations.}
While our joint 3D-multi-view approach achieves significant performance gains over previous state of the art in 3D semantic segmentation, there are still several important limitations.
Our approach operates on dense volumetric grids, which become quickly impractical for high resolutions; e.g., RGB-D scanning approaches typically produce reconstructions with sub-centimeter voxel resolution; sparse approaches, such as OctNet \cite{riegler2017OctNet}, might be a good remedy.
Additionally, we currently predict only the voxels of each column of a scene jointly, while each column is predicted independently, which can give rise to some label inconsistencies in the final predictions since different RGB views might be selected; note, however, that due to the convolutional nature of the 3D networks, the geometry remains spatially coherent.



\section{Conclusion and Future Work}

We presented \OURS{}, a joint 3D-multi-view approach built on the core idea of combining geometric and RGB features in a joint network architecture.
We show that our joint approach can achieve significantly better accuracy for semantic 3D scene segmentation.
In a series of evaluations, we carefully examine our design choices; for instance, we demonstrate that the 2D and 3D features complement each other rather than being redundant; we also show that our method can successfully take advantage of using several input views from an RGB-D sequence to gain higher coverage, thus resulting in better performance.
In the end, we are able to show results at more than {\bf 14\% higher classification accuracy} than the best existing 3D segmentation approach.
Overall, we believe that these improvements will open up new possibilities where not only the semantic content, but also the spatial 3D layout plays an important role.

For the future, we still see many open questions in this area. 
First, the 3D semantic segmentation problem is far from solved, and semantic instance segmentation in 3D is still at its infancy.
Second, there are many fundamental questions about the scene representation for realizing 3D convolutional neural networks, and how to handle mixed sparse-dense data representations.
And third, we also see tremendous potential for combining multi-modal features for generative tasks in 3D reconstruction, such as scan completion and texturing.


\section*{Acknowledgments}
This work was supported by a Google Research Grant, a Stanford Graduate Fellowship, and a TUM-IAS Rudolf M{\"o}{\ss}bauer Fellowship.

\bibliographystyle{splncs}
\bibliography{main}

\clearpage
\clearpage
\begin{appendix}


In this appendix, we provide additional quantitative results for 3D semantic segmentation using our \OURS{} method.
In particular, we use the Matterport3D~\cite{Matterport3D} benchmark for this purpose; see Sec.~\ref{sec:matterport}.
Note that the results on the ScanNet~\cite{dai2017scannet} and NYUv2~\cite{Silberman:ECCV12} datasets can be found in the main document.
Furthermore, in Sec.~\ref{sec:qualitative}, we visualize additional qualitative results.

\section{Evaluation on Matterport3D~\cite{Matterport3D}}
\label{sec:matterport}

Matterport3D provides $90$ building-scale RGB-D reconstructions, densely annotated similar to the ScanNet dataset annotations.
In Tab.~\ref{table:comparison_matterport}, we compare against state-of-the-art volumetric-based semantic 3D segmentation approaches (ScanNet~\cite{dai2017scannet} and ScanComplete~\cite{dai2018scancomplete}) on the Matterport3D~\cite{Matterport3D} dataset. 
Additionally, we evaluate the performance of our method over varying number of views in the ablation study in Tab.~\ref{table:ablation_matterport}.
Note that our final result improves over 10\% on compared to the best existing volumetric-based 3D semantic segmentation method.

\begin{table*}[!htb]
	\resizebox{\textwidth}{!}{%
		\centering
		\begin{tabular}{|l||c|c|c|c|c|c|c|c|c|c|c|c|c|c|c|c|c|c|c|c|c||c|}
			\hline
			& wall & floor & cab & bed & chair & sofa & table & door & wind & bkshf & pic & cntr & desk & curt & ceil & fridg & show & toil & sink & bath & other & avg \\ \hline
			ScanNet~\cite{dai2017scannet} & \textbf{80.1} & \textbf{97.3} & 46.5 & 75.6 & 67.7 & 34.2 & 35.7 & 45.6 & 15.6 & 4.6 & 4.2 & 36.0 & 9.2 & 3.3 & 96.5 & 0.0 & 0.0 & 53.4 & 24.5 & 40.4 & 8.5 & 37.1 \\ \hline
			ScanComplete~\cite{dai2018scancomplete} & 79.0 & 95.9 & 31.9 & 70.4 & 68.7 & 41.4 & 35.1 & 32.0 & 37.5 & \textbf{17.5} & 27.0 & \textbf{37.2} & \textbf{11.8} & 50.4 & \textbf{97.6} & 0.1 & 15.7 & \textbf{74.9} & 44.4 & 53.5 & \textbf{21.8} & 44.9 \\ \hline 
			\hline
			\textbf{\OURS{} (ours)} & 79.6 & 95.5 & \textbf{59.7} & \textbf{82.3} & \textbf{70.5} & \textbf{73.3} & \textbf{48.5} & \textbf{64.3} & \textbf{55.7} & 8.3 & \textbf{55.4} & 34.8 & 2.4 & \textbf{80.1} & 94.8 & \textbf{4.7} & \textbf{54.0} & 71.1 & \textbf{47.5} & \textbf{76.7} & 19.9 & \textbf{56.1} 		\\ \hline
		\end{tabular}
	}
	\vspace{0.1cm}
	\caption{Comparison of our final trained model (5 views, end-to-end) against the state-of-the-art volumetric-based semantic 3D segmentation methods on the Matterport3D dataset \cite{Matterport3D}.
		\vspace{-1.0cm}}
	\label{table:comparison_matterport}
\end{table*}

\begin{table*}[!htb]
	\resizebox{\textwidth}{!}{%
		\centering
		\begin{tabular}{|l||c|c|c|c|c|c|c|c|c|c|c|c|c|c|c|c|c|c|c|c|c||c|}
			\hline
			& wall & floor & cab & bed & chair & sofa & table & door & wind & bkshf & pic & cntr & desk & curt & ceil & fridg & show & toil & sink & bath & other & avg \\ \hline
			1-view & 76.5 & \textbf{97.0} & 57.6 & 75.9 &  \textbf{71.9} & 38.4 & 48.3 & 55.6 & 54.2 & 1.7 & 37.1 & \textbf{35.0} & 3.6 & 52.8 & \textbf{96.4} & 1.2 & 11.1 & 68.6 & 27.7 & 68.4 & 11.9 & 47.2 		\\ \hline
			3-view & 76.4 & 96.7 & 63.1 & \textbf{86.0} & 71.6 & 46.5 & 43.7 & \textbf{65.0} & 53.0 & \textbf{10.6} & \textbf{59.3} & 32.7 & \textbf{4.1} & 78.1 & 92.3 & 0.4 & 6.4 & \textbf{77.9} & 43.6 & 70.7 & \textbf{21.0} & 52.5 		\\ \hline
			\hline 
			\textbf{5-view} & \textbf{79.6} & 95.5 & \textbf{59.7} & 82.3 & 70.5 & \textbf{73.3} & \textbf{48.5} & 64.3 &\textbf{ 55.7} & 8.3 & 55.4 & 34.8 & 2.4 & \textbf{80.1} & 94.8 & \textbf{4.7} & \textbf{54.0} & 71.1 & \textbf{47.5} & \textbf{76.7} & 19.9 & \textbf{56.1} 		\\ \hline
		\end{tabular}
	}
	\vspace{0.1cm}
	\caption{Evaluation of 1-, 3-, and 5-view (end-to-end) variants of our approach on the Matterport3D~\cite{Matterport3D} test set. Note that each additional view improves the results by several percentage points, which confirms the findings of our ablation study on the ScanNet~\cite{dai2017scannet} dataset shown in the main document.
		\vspace{-1.0cm}}
	\label{table:ablation_matterport}
\end{table*}

\section{Additional Qualitative Results}
\label{sec:qualitative}

\begin{figure}[t!] \centering
	\includegraphics[width=0.92\linewidth]{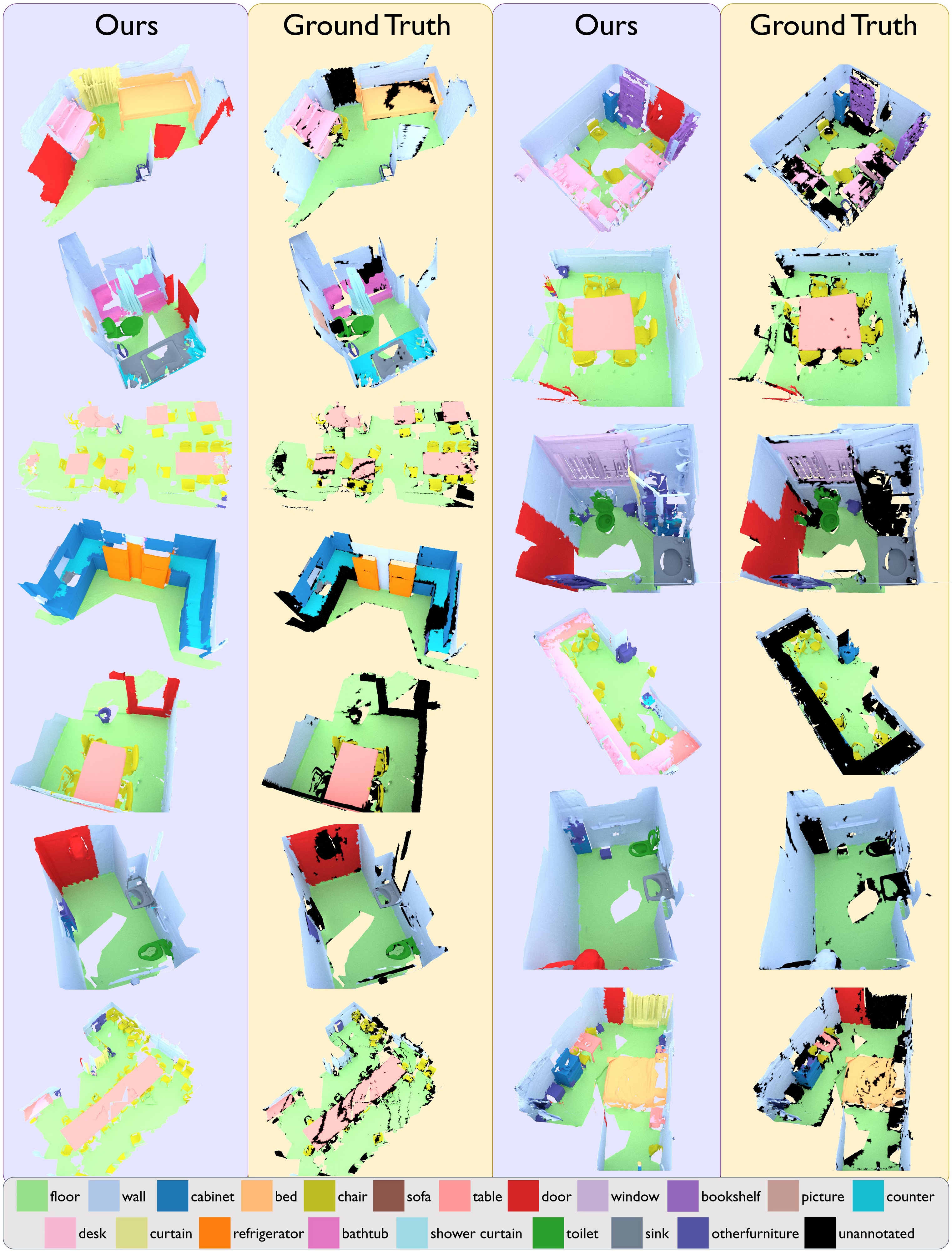}
	\caption{Additional qualitative semantic 3D segmentation results on the ScanNet~\cite{dai2017scannet} test set. Note that black denotes regions that are unannotated or contain labels not in the 20-label set.
		\vspace{-0.4cm}}
	\label{fig:gallery_scannet}
\end{figure}

\begin{figure}[t!] \centering
	\includegraphics[width=0.92\linewidth]{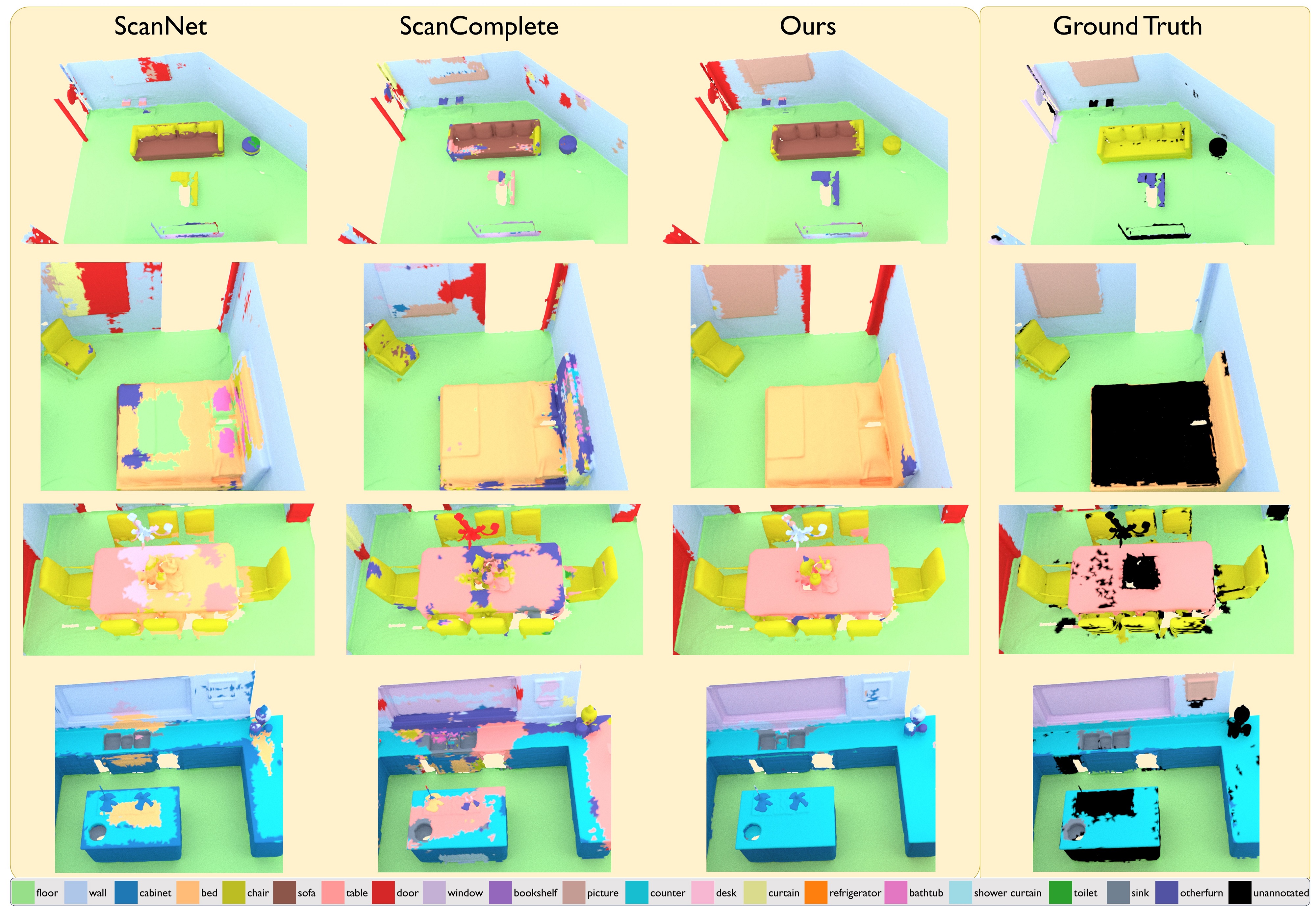}
	\caption{Qualitative semantic segmentation results on the Matterport3D~\cite{Matterport3D} test set. We compare with the 3D-based approaches of ScanNet~\cite{dai2017scannet} and ScanComplete~\cite{dai2018scancomplete} Note that black denotes regions that are unannotated or contain labels not in the label set.
		\vspace{-0.4cm}}
	\label{fig:comparison_matterport}
\end{figure}

In Fig.~\ref{fig:gallery_scannet}, we show additional qualtitative results using our \OURS{} approach.
We additionally show a qualitative comparison to the volumetric semantic segmentation approaches of ScanNet~\cite{dai2017scannet} and ScanComplete~\cite{dai2018scancomplete} on the Matterport3D~\cite{Matterport3D} dataset in Fig.~\ref{fig:comparison_matterport}.

\end{appendix}

\end{document}